\documentclass[10pt,twocolumn,letterpaper]{article}

\usepackage{cvpr}              

\usepackage{graphicx}
\usepackage{amsmath}
\usepackage{amssymb}
\usepackage{booktabs}
\usepackage{multirow}
\usepackage[accsupp]{axessibility} 

\usepackage[pagebackref,breaklinks,colorlinks]{hyperref}

\usepackage[capitalize]{cleveref}
\crefname{section}{Sec.}{Secs.}
\Crefname{section}{Section}{Sections}
\Crefname{table}{Table}{Tables}
\crefname{table}{Tab.}{Tabs.}

\def\cvprPaperID{733} 
\def\confName{CVPR}
\def\confYear{2023}

\begin{document}

\title{Parametric Implicit Face Representation for Audio-Driven Facial Reenactment
}


\author{Ricong Huang
$^{1}$ \quad Peiwen Lai$^{1}$ \quad Yipeng Qin$^{2}$ \quad Guanbin Li$^{1}$\thanks{Corresponding author is Guanbin Li.} \\
$^1${School of Computer Science and Engineering, Sun Yat-sen University} \quad $^2${Cardiff University} \\
{\tt\small \{huangrc3, laipw5\}@mail2.sysu.edu.cn, qiny16@cardiff.ac.uk, liguanbin@mail.sysu.edu.cn}
}

\maketitle

\begin{abstract}
Audio-driven facial reenactment is a crucial technique that has a range of applications in film-making, virtual avatars and video conferences. Existing works either employ explicit intermediate face representations (\eg, 2D facial landmarks or 3D face models) or implicit ones (\eg, Neural Radiance Fields), thus suffering from the trade-offs between interpretability and expressive power, hence between controllability and quality of the results. In this work, we break these trade-offs with our novel parametric implicit face representation and propose a novel audio-driven facial reenactment framework that is both controllable and can generate high-quality talking heads. Specifically, our parametric implicit representation parameterizes the implicit representation with interpretable parameters of 3D face models, thereby taking the best of both explicit and implicit methods. In addition, we propose several new techniques to improve the three components of our framework, including i) incorporating contextual information into the audio-to-expression parameters encoding; ii) using conditional image synthesis to parameterize the implicit representation and implementing it with an innovative tri-plane structure for efficient learning; iii) formulating facial reenactment as a conditional image inpainting problem and proposing a novel data augmentation technique to improve model generalizability. Extensive experiments demonstrate that our method can generate more realistic results than previous methods with greater fidelity to the identities and talking styles of speakers. 
\end{abstract}

\begin{figure}[t]
  \centering
  \includegraphics[width=\linewidth]{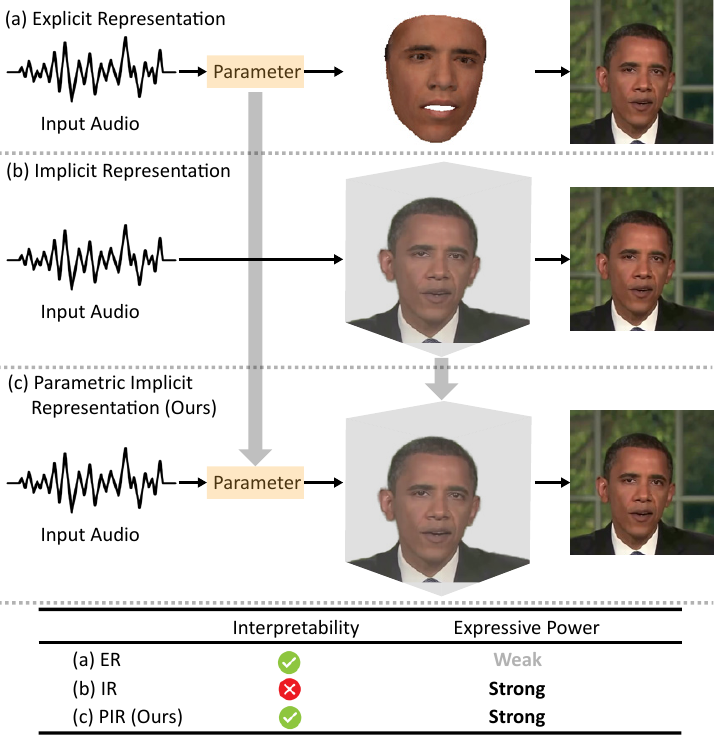}
  \caption{Comparison between previous explicit, implicit representations and our parametric implicit representation (PIR). (a) Explicit representations (\eg, 3D face models) have interpretable parameters but lack expressive power.
  (b) Implicit representations (\eg, NeRF) have strong expressive power but are not interpretable.
  (c) Our PIR takes the best of both approaches and is both interpretable and expressive, thus paving the way for controllable and high-quality audio-driven facial reenactment. }
  \label{fig:teaser}
\end{figure}

\section{Introduction}
\label{sec:intro}

Audio-driven facial reenactment, also known as audio-driven talking head generation or synthesis, plays an important role in various applications, such as digital human, film-making and virtual video conference. It is a challenging cross-modal task from audio to visual face, which requires the generated talking heads to be photo-realistic and have lip movements synchronized with the input audio.

According to the intermediate face representations, existing facial reenactment methods can be roughly classified into two categories: explicit and implicit methods.
Between them, explicit methods~\cite{suwajanakorn2017synthesizing, chen2019hierarchical, zhou2020makelttalk, lu2021live, xie2021towards, thies2020neural, yi2020audio, zhang2021facial, song2022everybody} exploit relatively sophisticated 2D (\eg, 2D facial landmarks~\cite{suwajanakorn2017synthesizing, chen2019hierarchical, zhou2020makelttalk, lu2021live, xie2021towards}) or 3D (\eg, 3D Morphable Model~\cite{thies2020neural, yi2020audio, zhang2021facial, song2022everybody}) parametric face models to reconstruct 2D or 3D faces, and map them to photo-realistic faces with a rendering network such as the Generative Adversarial Networks (GANs)~\cite{wang2018high, yu2019free}.
Their distinct advantage is the controllability (\eg, expressions) resulting from their interpretable facial parameters. 
However, despite this advantage, the parametric face models used in explicit methods are often sparse and have very limited expressive power, which inevitably sacrifices the quality of synthesized faces (\eg, the inaccurate lip movements and blurry mouth caused by the missing teeth area in 3D face models).
In contrast, implicit methods~\cite{zhou2019talking, prajwal2020lip, sun2021speech2talking, zhou2021pose, liang2022expressive, guo2021ad, liu2022semantic, shen2022learning} use implicit 2D or 3D representations that are more expressive and can generate more realistic faces.
For example, Neural Radiance Fields (NeRF) based methods \cite{guo2021ad, liu2022semantic, shen2022learning} are one of the more representative implicit methods that use NeRF to represent the 3D scenes of talking heads.
Although being more expressive and producing higher-quality results, 
implicit methods are not interpretable and lose the controllability of the synthesis process, thus requiring model re-training to change its target person. 
As a result, the explicit and implicit methods mentioned above form a trade-off between the {\it interpretability} and {\it expressive power} of intermediate face representations, while a representation that is both interpretable and expressive remains an open problem.

In this work, we break the above trade-off by proposing a parametric implicit representation that is both interpretable and expressive, paving the way for controllable and high-quality audio-driven facial reenactment.
Specifically, we propose to parameterize implicit face representations with the interpretable parameters of the 3D Morphable Model (3DMM)~\cite{deng2019accurate} using a conditional image synthesis paradigm. In our parametric implicit representation, the 3DMM parameters offer interpretability and the implicit representation offers strong expressive power, which take the best of both explicit and implicit methods (\cref{fig:teaser}). 
To implement our idea, we propose a novel framework consisting of three components: i) contextual audio to expression (parameters) encoding; ii) implicit representation parameterization; iii) rendering with parametric implicit representation.
Among them, our {\it contextual audio to expression encoding} component employs a transformer-based encoder architecture to capture the long-term context of an input audio, making the resulting talking heads more consistent and natural-looking; our {\it implicit representation parameterization} component uses a novel conditional image synthesis approach for the parameterization, and innovatively employs a tri-plane based generator offered by EG3D~\cite{chan2022efficient} to learn the implicit representation in a computationally efficient way; our {\it rendering with parametric implicit representation} component formulates face reenactment as an image inpainting problem conditioned on the parametric implicit representation to achieve a consistent and natural-looking ``blending'' of the head and torso of a target person.
In addition, we observe that the model slightly overfits to the training data consisting of paired audio and video, causing jitters in the resulting talking heads whose lip movements are required to be synchronized with unseen input audio.
To help our model generalize better and produce more stable results, we further propose a simple yet effective data augmentation strategy for our rendering component.

In summary, our main contributions include:

\begin{itemize}
    \item We propose an innovative audio-driven facial reenactment framework based on our novel parametric implicit representation, which breaks the previous trade-off between interpretability and expressive power, paving the way for controllable and high-quality audio-driven facial reenactment.
    
    \item We propose several new techniques to improve the three components of our innovative framework, including: i) employing a transformer-based encoder architecture to incorporate contextual information into the audio to expression (parameters) encoding; ii) using a novel conditional image synthesis approach for the parameterization of implicit representation, which is implemented with an innovative tri-plane based generator \cite{chan2022efficient} for efficient learning; iii) formulating facial reenactment as a conditional image inpainting problem for natural ``blending'' of head and torso, and proposing a simple yet effective data augmentation technique to improve model generalizability.
    
    \item Extensive experiments show that our method can generate high-fidelity talking head videos and outperforms state-of-the-art methods in both objective evaluations and user studies.
\end{itemize}

\section{Related work}
\label{sec:related}

Given a video of a target person and an (unpaired) audio, audio-driven facial reenactment aims to synthesize a novel video of the target person whose lip movement is synchronized with the given audio.
Most existing talking head generation methods can be roughly classified into two categories: explicit methods and implicit methods, according to their intermediate face representations.

\vspace{2mm}
\noindent \textbf{Explicit Methods.}
Explicit methods use parametric face models as intermediate face representations.
Depending on the type of parametric face models used, explicit methods can be further divided into two categories: 2D-based and 3D-based.
Between them, 2D-based methods use 2D parametric face models like 2D facial landmarks \cite{suwajanakorn2017synthesizing, chen2019hierarchical, zhou2020makelttalk, lu2021live, xie2021towards}, and map the input audio to them. 
These 2D landmarks are then fed into generative adversarial networks (GANs) to synthesize photo-realistic faces. 
For example, Chen \etal \cite{chen2019hierarchical} propose an adjustable pixel-wise loss to guide the network to focus on audiovisual-correlated facial landmarks. 
Xie \etal \cite{xie2021towards} predict the facial landmarks in the mouth area with the input audio and then change the lip movement of video frames to match the predicted landmarks. 
In contrast, 3D-based methods use more expressive 3D parametric face models (\eg, the 3D Morphable Models (3DMM) \cite{blanz1999morphable, paysan20093d} and FLAME \cite{li2017learning}) and map the audio to the expression parameters of the models. 
These expression parameters, together with those extracted from the video frames, are used to reconstruct explicit 3D face shapes that will be fed into a rendering network to synthesize new talking head videos.
For example, Thies \etal \cite{thies2020neural} encode the audio into a general audio-expression space and learn a person-specific expression basis to reconstruct the intermediate 3D model. 
Zhang \etal \cite{zhang2021facial} leverage the context in audio to model implicit attributes like eye blinking and head pose, extending the attribute control of the face model. 
Song \etal \cite{song2022everybody} remove identity information in audio to improve the quality of expression parameters. And they exploit the use of expression and landmarks from video frames to supervise the reconstructed facial mesh. Despite being interpretable, all parametric face models used in explicit methods are sparse compared to image pixels and cannot capture facial details (\eg, the missing areas like teeth in 3D face models). 



\vspace{2mm}
\noindent \textbf{Implicit Methods.} 
Some works achieve the audio-to-face transition directly through the Generative Adversarial Networks (GANs) \cite{zhou2019talking, prajwal2020lip, sun2021speech2talking, zhou2021pose, liang2022expressive, ye2022audio}. Prajwal \etal \cite{prajwal2020lip} employ a powerful lip-sync discriminator to detect lip-sync errors, forcing the generator to extract more expressive representations from the input audio. Zhou \etal \cite{zhou2021pose} devise an implicit pose code to achieve free pose control and enhance the audio representation by contrastive learning in a non-identity space. 
Recently, some other implicit methods use Neural Radiance Fields (NeRF) \cite{mildenhall2021nerf} as the intermediate representation \cite{guo2021ad, liu2022semantic, zhou2022dialoguenerf, shen2022learning}, which models the 3D scene of a talking head with a fully-connected network and volume rendering techniques.
For example, Guo \etal \cite{guo2021ad} employ two individual sets of NeRF to synthesize the talking head and torso of a portrait respectively. 
Liu \etal \cite{liu2022semantic} leverage the semantic information in video frames to guide NeRF to concentrate on the hard-to-learn area like mouth and eyes. Shen \etal \cite{shen2022learning} introduce audio conditions to warp the face to the query space, which is applied in the fine-tuning of the facial radiance field for few-shot synthesis. 
Although implicit methods have more expressive representations and produce higher-quality videos, they are less interpretable and lose the controllability of the synthesis process, thus requiring model re-training to change its target person.


\vspace{2mm}
In this work, we propose a novel framework that takes the best of both explicit and implicit methods. Specifically, we exploit the interpretable parameter space of 3D parametric face models, but map them to implicit face representations instead of reconstructing 3D face models. 
In this way, we obtain a representation that is both expressive and interpretable, thus paving the way for controllable and high-quality audio-driven facial reenactment.
\begin{figure*}[t]
  \centering
   \includegraphics[width=\linewidth]{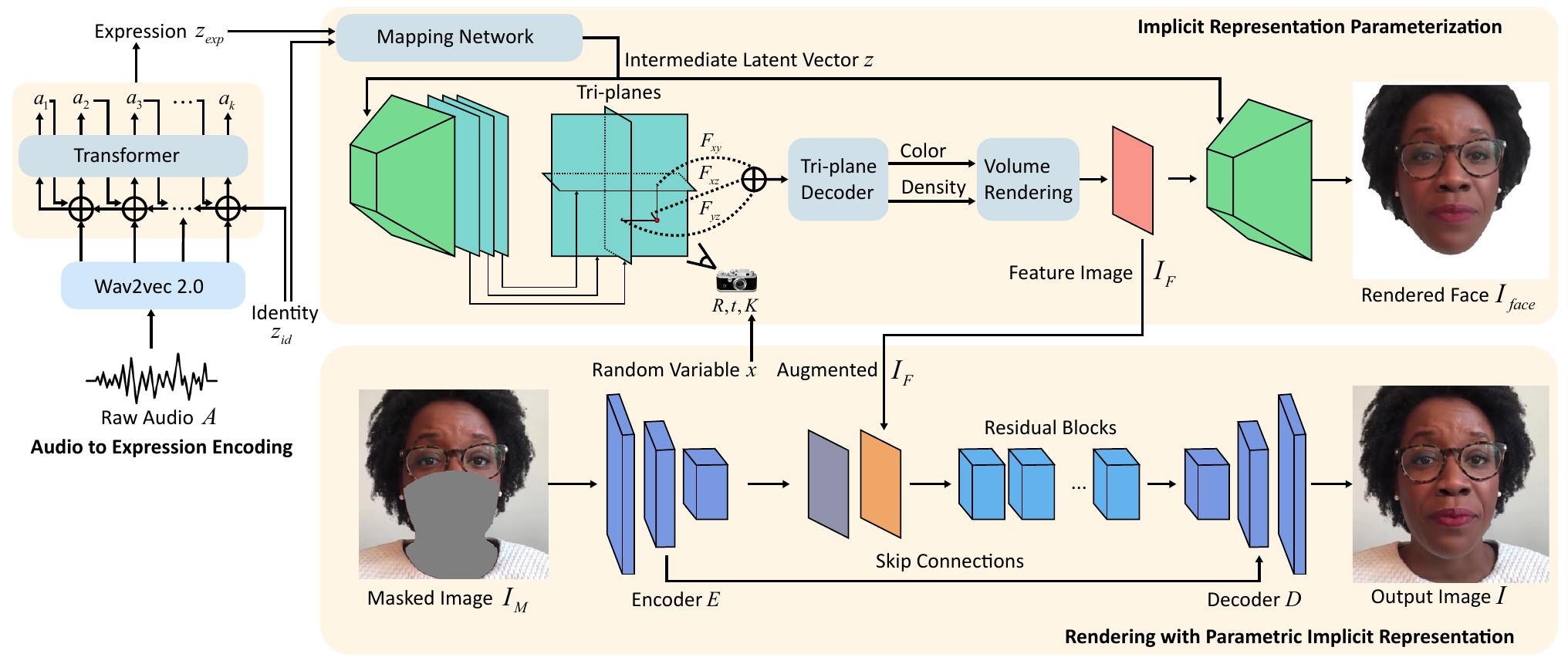}

   \caption{Overview of our framework. Our framework contains three components: i) contextual audio to expression (parameters) encoding; ii) implicit representation parameterization; iii) rendering with parametric implicit representation (PIR). With a given identity parameter $z_{id}$, raw audio $A$ is mapped to the expression parameters $z_{exp}$ which captures long-term context from $\{a_i | i=1,2,...,k-1\}$ through a transformer network. 
   Our implicit representation parameterization component first maps $z_{exp}$ and $z_{id}$ into a latent vector $z$ to condition the generation of tri-plane feature maps, and then sample them with pose parameters $R$, $t$, $K$ to obtain pose-conditioned features. These pose-conditioned features are processed with a lightweight decoder and a volume rendering module to produce our PIR $I_F$. 
   An upsampling network is used to generate the face image $I_{face}$ from $I_F$. 
   Our rendering with PIR component generates output image $I$ by formulating it as an inpainting task conditioned on a masked image $I_M$ and $I_F$. }
   \label{fig:framework}
\end{figure*}

\section{Methodology}

As \cref{fig:framework} shows, our framework consists of three components: contextual audio to expression (parameters) encoding (\cref{sub:audio2expression}), implicit representation parameterization (\cref{sub:faceBranch}) and rendering with parametric implicit representation (\cref{sub:renderBranch}).
Given an input raw audio $A$, our {\it contextual audio to expression encoding} component maps $A$ to its corresponding expression parameter $z_{exp}$ in the same format as used in 3D Morphable Model (3DMM). $z_{exp}$, together with the identity parameter $z_{id}$ and the pose parameter (rotation $R$, translation $t$ and camera intrinsic matrix $K$, which are used as the camera pose) extracted by 3DMM, constitute the facial parameters and are mapped to the implicit representation of a reenacted face $I_F$ by our {\it implicit representation parameterization} component.
Finally, our {\it rendering with parametric implicit representation (PIR)} component formulates facial reenactment as a conditional inpainting task and renders the reenacted image with $I_F$ as the condition.
\subsection{Contextual Audio to Expression Encoding}
\label{sub:audio2expression}

As \cref{fig:framework} shows, unlike previous implicit methods that train the audio encoder in an end-to-end fashion~\cite{guo2021ad, liu2022semantic}, we explicitly supervise its training with expression parameters extracted by 3D Morphable Model (3DMM). The rationale behind our choice is that audio has little to do with a person's identity or pose but mainly his/her expression (\eg, lip movement).
Specifically, given a raw audio $A$, we first extract its preliminary feature using wav2vec 2.0 \cite{baevski2020wav2vec}, a self-supervised pre-trained speech model that facilitates accurate lip movement through the abundant phoneme information it learned from a large-scale corpus of unlabeled speech.
Then, we feed this feature along with the identity parameters extracted by 3DMM into a transformer-based audio feature extractor network \cite{fan2022faceformer} which encodes it to the expression parameters $a_k$ of the $k$-th video frame. 
Thanks to the transformer architecture, $a_k$ is dependent on the expression parameters of previous frames $\{a_i | i=1,2,\dots,k-1\}$ and effectively captures the context of the audio.
In addition, our method separates audio encoding as a stand-alone and light-weight subtask, which can make the most of the computational resources and capture much longer-term dependency (\ie using longer input sequences), resulting in more consistent and natural-looking videos.


\subsection{Implicit Representation Parameterization}
\label{sub:faceBranch}

Unlike previous methods that reconstruct 3D face shapes with the extracted facial parameters~\cite{zhang2021facial, thies2020neural} and convert them to videos, we map such facial parameters to an implicit representation $I_F$ and use $I_F$ to condition the video synthesis.
In this way, our framework takes the best of both explicit and implicit face representation approaches as 
i) it makes good use of the interpretability of the facial parameter space that facilitates controllability of the synthesis process; 
ii) it captures more realistic facial features with the high expressive power of the implicit representation; iii) it avoids the unnecessary introduction of facial priors that are inconsistent with ground truth when performing 3D face reconstruction from sparse facial parameters.

As \cref{fig:framework} shows, we implement the mapping between facial parameters and its implicit representation using a EG3D\cite{chan2022efficient} generator.
Specifically, our facial parameters consist of three components: identity, expression and pose.
For identity and expression, we concatenate them and employ a simple mapping network to map them to an intermediate latent vector $z$. 
For pose, we represent it with the camera pose $R, t$ and intrinsic matrix $K$, and use it to query the 3D positions using the tri-plane structure. 
Following \cite{chan2022efficient}, we feed $z$ to the EG3D generator as both input and condition vector, and $R, t, K$ to it as the camera pose, and obtain $I_F$ as an implicit representation of the input facial parameters.
Note that we use face reconstruction as a proxy task ($I_{face}$ denotes the reconstructed face) for the training and discard the decoder in the testing stage.


\vspace{2mm}
\noindent \textbf{Remark.} 
We use EG3D~\cite{chan2022efficient} rather than NeRF~\cite{guo2021ad, yang2022psnerf} as EG3D is a computationally efficient and expressive architecture that supports the generation of high-resolution images in real time and greatly preserves 3D structure. 


\subsection{Rendering with PIR}
\label{sub:renderBranch}

As mentioned above, although the implicit representation $I_F$ carries realistic facial features, its sparse input (\ie, facial parameters) cannot capture the fine details of the input video.
To this end, we formulate facial reenactment as a video inpainting problem conditioned on the implicit representation $I_F$.
Specifically, as shown in \cref{fig:framework}, given a masked video frame $I_M$, we first use an encoder $E$ to extract its feature maps with the same resolution as $I_F$. Then, we concatenate them with $I_F$ and feed the concatenated feature maps to the decoder $D$ to generate the reenactment image $I$.
Skip connections are added between corresponding intermediate layers of $E$ and $D$. 

\begin{figure}[t]
  \centering
   \includegraphics[width=\linewidth]{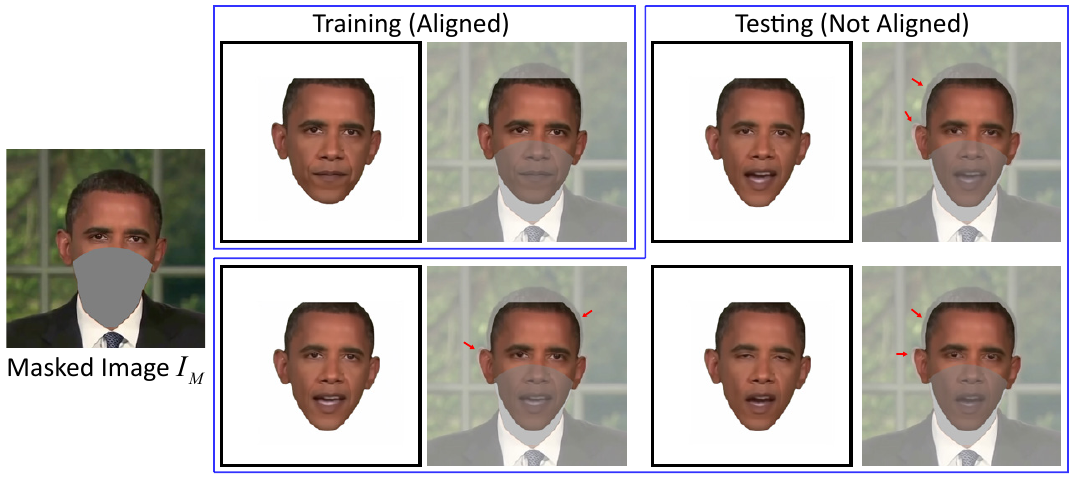}
    \caption{Jitters caused by training-inference mismatch. 
    In training, $I_{face}$ is well aligned with the masked image $I_M$, which is not the case in testing, thus causing jitters. 
    }
   \label{fig:jitters}
\end{figure}
\begin{figure}[t]
  \centering
   \includegraphics[width=\linewidth]{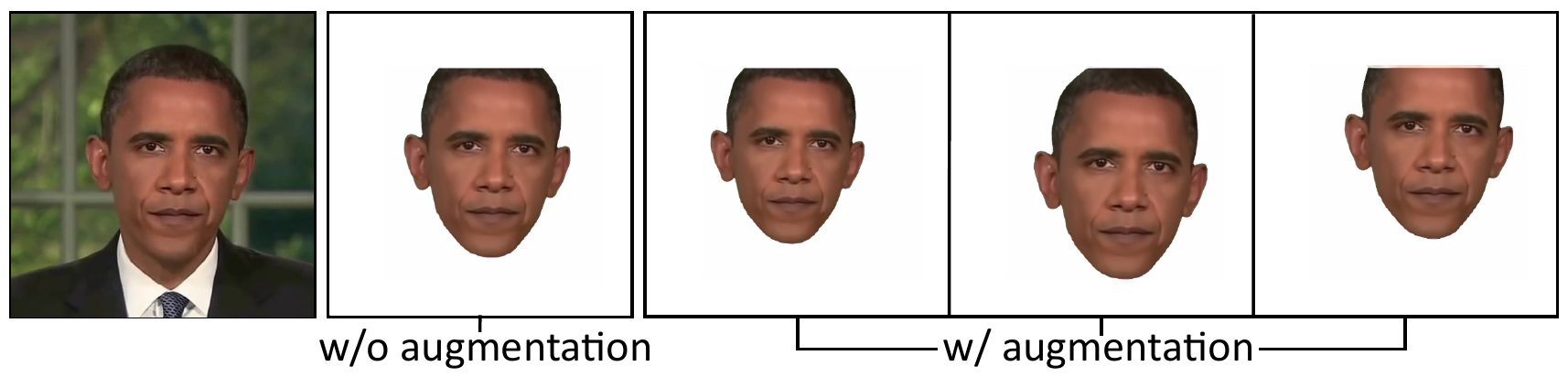}

   \caption{Data augmentation for jitter reduction. We augment the camera intrinsic matrix $K$ to randomly scale and shift $I_F$ to simulate potential training-inference mismatches during training.}
   \label{fig:intrinsic}
\end{figure}

\vspace{2mm}
\noindent \textbf{Jitter Reduction.}
Although effective, the proposed rendering method is trained with paired audio and video data, which is not the case during testing.
In our experiment (\cref{fig:jitters}), we observed that new audios may cause slight offsets and deformations of $I_{face}$, leading to jitters in the resulting videos. 
To reduce such jitters, we propose a simple yet effective data augmentation strategy. Specifically, we perturb the camera intrinsic matrix $K$ with random variables $x_1, x_2, x_3 \sim U(-s, s)$ when training the rendering network:
\begin{equation}
    K = \begin{bmatrix}
        f_x \times (1+\frac{x_1}{2u_0}) & 0 & u_0 + x_2 \\
        0 & f_y \times (1+\frac{x_1}{2v_0}) & v_0 + x_3 \\
        0 & 0 & 1
        \end{bmatrix},
\end{equation}
where $f_x$ and $f_y$ represent focal length in terms of pixels and $u_0$ and $v_0$ represent the principal point. 
This augmentation simulates the scaling and shifting of $I_F$ and thus $I_{face}$ (\cref{fig:intrinsic}), allowing the rendering network to learn from them and reduce jitters.

\subsection{Training and Loss Functions}
\label{sub:loss}

Among the three components of our framework, {\it audio to expression} is a self-contained subtask and can be trained independently, which makes the most of the given computational resources and produces more consistent and natural-looking videos; {\it rendering} relies on the result of {\it implicit representation parameterization} and is trained afterwards.
All of them are trained with short video clips of target persons, including paired audio track and video sequences.

\vspace{2mm}
\noindent \textbf{Audio to Expression.}
Let $a_i$ be the output feature where $i=1,2,...,k$ be the $i$-th frame of the video, and $z_{exp, i}$ be the expression parameter extracted by 3DMM, we train our audio encoder by minimizing the Mean Squared Error (MSE) between them as:
\begin{equation}
    L_{audio} = \sum_{i=1}^k||a_i - z_{exp, i}||^2.
\end{equation}

\vspace{2mm}
\noindent \textbf{Implicit Representation Parameterization.}
We train the mapping between input facial parameters and the implicit representation of reenacted face with a weighted sum of a photometric loss and a perceptual loss \cite{johnson2016perceptual}
\begin{equation}
    \begin{aligned}
        L_{face} &= w_1||M_h \odot (I_{face} - I_{GT})||^2 + \\
        & w_2\sum_{i}||\phi_i(I_{face}) - \phi_i(M_h \odot I_{GT})||^2
    \end{aligned},
    \label{equ:L_face}
\end{equation}
where $M_h$ is the head mask, $\phi_i(*)$ denotes the activation of the $i$-th layer in VGG16 \cite{simonyan2014very}, $\odot$ denotes element-wise product operator, $w_1$ and $w_2$ are the weighting factors. 

\vspace{2mm}
\noindent \textbf{Rendering with PIR.}
To maximize the quality of generated image, we train our rendering network with:
\begin{equation}
    L_{render} = L_{render}^{rec} + L_{render}^{FM} + L_{render}^{GAN}
\end{equation}
where $L_{render}^{rec}$ denotes a reconstruction loss consisting of a weighted sum of a photometric loss and a perceptual loss:
\begin{equation}
    L_{render}^{rec} = w_3||I - I_{GT}||^2
    + w_4\sum_{i}||\phi_i(I) - \phi_i(I_{GT})||^2.
    \label{equ:L_render_rec}
\end{equation}
Let $\{D_k | k=1, 2, 3\}$ be a multi-scale discriminator \cite{wang2018high}, $L_{render}^{FM}$ denotes a feature matching loss and $L_{render}^{GAN}$ denotes a GAN adversarial loss:
\begin{equation}
    \begin{aligned}
        &L_{render}^{FM} = \sum_{i=1}^T\frac{1}{N_i}[||D_k^{(i)}(I) - D_k^{(i)}(I_{GT})||_1] \\
        &L_{render}^{GAN} = \sum_{k} \log D_k(I_{GT}) + \log (1-D_k(I))
    \end{aligned},
\end{equation}
where $T$ is the total number of layers and $N_i$ denotes the number of elements in each layer. Note that $L_{render}^{GAN}$ is optimized in a minimax manner as those in GAN training.

\begin{table*}[t]
  \centering
  \begin{tabular}{lccccc|cc|cc}
    \toprule
    \multirow{2}*{Method} & \multicolumn{5}{c}{HDTF} & \multicolumn{2}{c}{Testset 1} & \multicolumn{2}{c}{Testset 2} \\
    \cmidrule(r){2-6}\cmidrule(r){7-8}\cmidrule(r){9-10}
    & SSIM$\uparrow$ & PSNR$\uparrow$ & CPBD$\uparrow$ & LMD$\downarrow$ & AVConf$\uparrow$ & LMD$\downarrow$ & AVConf$\uparrow$ & LMD$\downarrow$ & AVConf$\uparrow$ \\
    \midrule
    Ground Truth & 1 & N/A & 0.344 & 0 & 8.839 & 0 & 8.407 & 0 & 9.315 \\
    \midrule
    ATVG \cite{chen2019hierarchical} & \underline{0.829} & 20.540 & 0.078 & 9.645 & 4.848 & 8.784 & 5.232 & 10.445 & 4.903 \\
    Wav2Lip \cite{prajwal2020lip} & 0.729 & 20.352 & {\bf 0.317} & \underline{4.279} & {\bf 7.812} & 4.836 & {\bf 7.554} & 4.297 & {\bf 7.682} \\
    MakeitTalk \cite{zhou2020makelttalk} & 0.698 & 19.956 & 0.075 & 4.940 & 3.972 & 4.939 & 4.172 & 5.064 & 3.467 \\
    PC-AVS \cite{zhou2021pose} & 0.738 & \underline{21.078} & 0.096 & 5.199 & \underline{7.392} & 6.678 & \underline{6.742} & 4.091 & 6.858 \\
    AD-NeRF \cite{guo2021ad} & - & - & - & - & - & \underline{4.691} & 4.236 & - & - \\
    FACIAL \cite{zhang2021facial} & - & - & - & - & - & - & - & \underline{2.675} & 6.045 \\
    \midrule
    Ours & {\bf 0.970} & {\bf 36.711} & \underline{0.305} & {\bf 1.794} & 7.233 & {\bf 2.116} & 6.133 & {\bf 1.899} & \underline{7.655} \\
    \bottomrule
  \end{tabular}
    \caption{Quantitative comparisons with existing state-of-the-art methods. Since AD-NeRF \cite{guo2021ad} and FACIAL \cite{zhang2021facial} do not provide pretrained models on the HDTF dataset, we only compare with them on Testset 1 and 2. {\bf Bold}: best results; \underline{Underline}: second-best results.}
  \label{tab:quantitative}
\end{table*}

\section{Experiments}
\subsection{Experimental Settings}
\noindent{\bf Dataset.} 
To achieve high-quality audio-driven facial reenactment, we follow \cite{guo2021ad, liu2022semantic} and conduct experiments on three datasets, 
the HDTF \cite{zhang2021flow}, Testset 1 \cite{guo2021ad}, Testset 2 \cite{zhang2021facial}.
For HDTF, we selected 8 videos (corresponding to 8 subjects) from it. For Testset 1 and 2, we use the sole video released by the authors for each of them. The time span of these videos is 3-6 minutes. 
For the test set consisting of unpaired and gender-balanced audio clips, we select 3 from HDTF and collect another 2 Obama audios online.
Please note that many previous datasets (\eg, LRW \cite{chung2016lip}, Voxceleb1 \cite{nagrani2017voxceleb} and Voxceleb2 \cite{chung2018voxceleb2}) are not suitable as they either have low image quality or consist of many short (a few seconds) video clips of different speakers (\eg, GRID \cite{cooke2006audio} and MEAD \cite{wang2020mead}), which hinders the generation of high-quality videos and the capture of long-term audio context. 

\vspace{2mm}
\noindent{\bf Data Preprocessing.} 
Before use, all videos are extracted at 25 frames per second (FPS) and the synchronized audio waves are sampled as $16K$ Hz frequency.
We crop the videos to center the faces and resize them to the resolution of $512 \times 512$. For each video, we extract the identity, expression and pose parameters of each frame using 3DMM \cite{deng2019accurate}. 
We obtain the (mean) identity parameters of a target person by averaging those of the same person over all frames in a video. 
The estimated head poses are represented as camera poses. An off-the-shelf segmentation method \cite{yu2018bisenet} is used to obtain the parsing map (\eg, head mask) of each frame. 

\vspace{2mm}
\noindent{\bf Evaluation Settings.} 
To facilitate an objective evaluation of lip movement accuracy and image quality, we first evaluate our method under self-reenactment setting on all three datasets, with the last 25s of each original video clip being used as the ground truth test data. 
To evaluate the performance of our method across identities (where there are no ground truths), we pair the videos in Testset 1 and 2 with audios of different identities for reenactment and use SyncNet~\cite{chung2016out} to assess the quality of synchronization.

\subsection{Implementation details}
We implement our framework in PyTorch \cite{paszke2019pytorch} with an Adam optimizer \cite{kingma2014adam}. 
We train our contextual audio to expression component for 120 epochs with the expression parameters extracted by 3DMM, a context length of $k=100$ (each frame lasts for 0.04s), and pretrained wav2vec 2.0 weights.
Thanks to its well-defined output, we train our audio to expression component simultaneously with the other two components. 
Our implicit representation paramerterization component is trained for 50 epochs using the expression parameters extracted by 3DMM as input and the rendered face $I_{face}$ as output.
The resolution of the resulting parametric implicit representation $I_F$ is $32\times 64 \times 64$.
The three tri-plane features have the resolution of $32\times 256 \times 256$ and that of output image $I_{face}$ is $3\times 512\times 512$. 
Our rendering with parametric implicit representation component is trained for another 50 epochs with $s=3$ for jitter reduction and the augmented $I_F$ whose resolution is $32\times 32 \times 32$. 
We use $w_1=w_2=w_3=w_4=1$ for \cref{equ:L_face,equ:L_render_rec}.

\subsection{Comparison with the State-of-the-arts 
}
\subsubsection{Quantitative Evaluation}

We quantitatively compare our method with SOTAs using the following metrics:
\begin{itemize}
    \item \noindent{\bf Lip Movement Accuracy:} We use the Landmark Distance (LMD) \cite{chen2018lip} to evaluate lip movement accuracy.
    \item \noindent{\bf Lip-sync:} We measure lip synchronization errors with the Audio-Visual Confidence (AVConf) score calculated by SyncNet \cite{chung2016out}. 
    \item \noindent{\bf Sharpness:} We measure frame sharpness with the perceptual-based no-reference objective image sharpness metric (CPBD) \cite{narvekar2011no}. 
    \item \noindent{\bf Image Quality:} We assess the quality of synthesized video frames by comparing them with the ground truth using Peak Signal to Noise Ratio (PSNR) and Structure Similarity Index Measure (SSIM) \cite{wang2004image}.
\end{itemize}

\begin{figure*}[t]
  \centering
   \includegraphics[width=\linewidth]{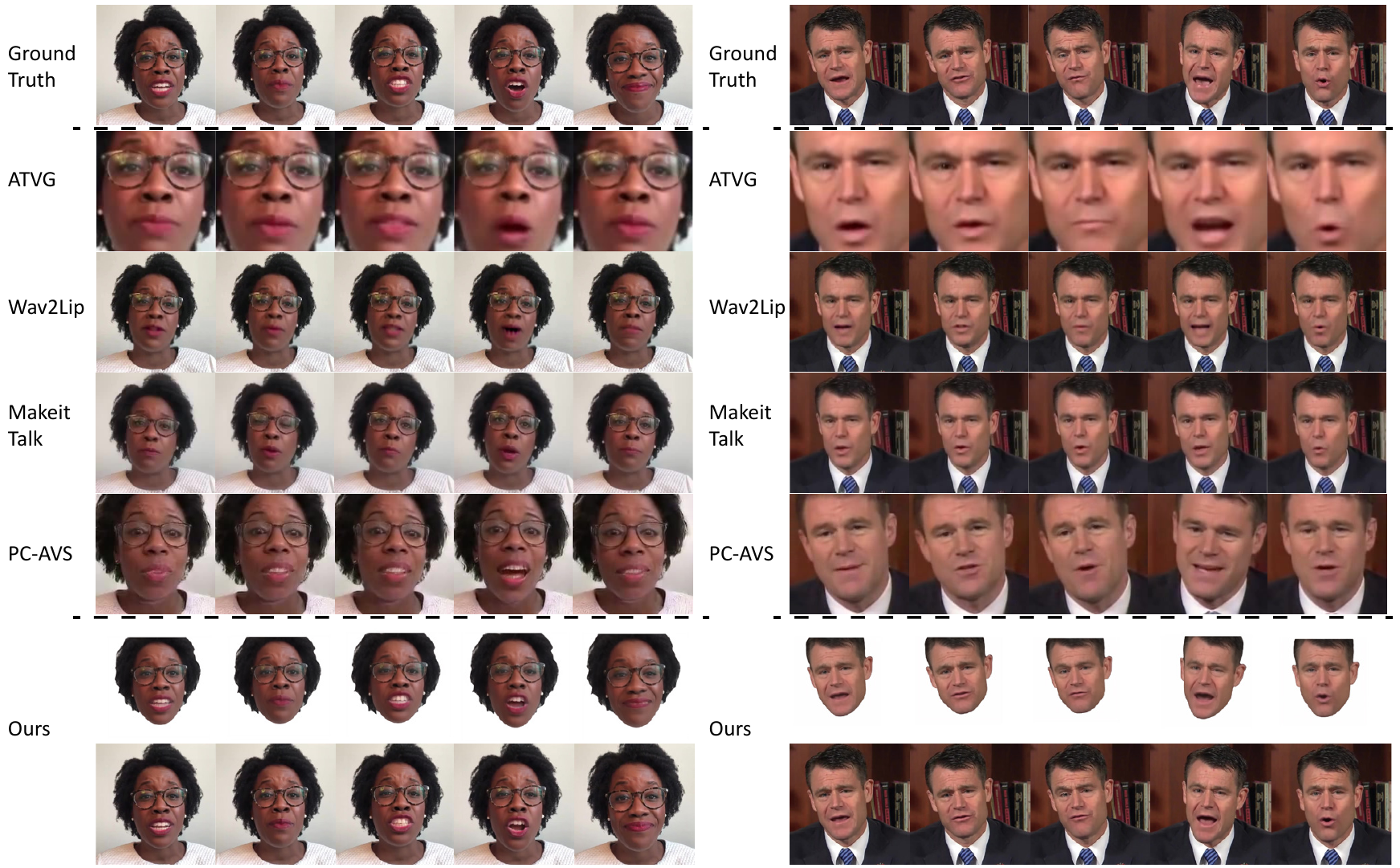}

\caption{Qualitative comparison with ATVG \cite{chen2019hierarchical}, Wav2Lip \cite{prajwal2020lip}, MakeitTalk \cite{zhou2020makelttalk} and PC-AVS \cite{zhou2021pose}. Row 1 (Ground Truth): the corresponding video frames to the input audio. Last two rows (Ours): rendered face $I_{face}$ and output image $I$ (\cref{fig:framework}), respectively. 
The portraits generated by our method have highly synchronized lip movements and fine facial details (\eg, teeth and wrinkles) that preserve the target person's identity well, thus outperforming all previous methods.
}
   \label{fig:qualitative}
\end{figure*}

As \cref{tab:quantitative} shows, our method achieves the best or second-best performance for most of the evaluation metrics. 
For the exceptions, Wav2Lip achieves a higher CPBD score on HDTF but sacrifices the visual quality (blurry mouths with obvious artifacts in \cref{fig:qualitative}) as it only edits the mouth region of the reference images 
with the remaining part unchanged.
In addition, it is trained using a pretrained lip-sync discriminator similar to SyncNet, which ``tricks'' SyncNet to produce the highest AVConf scores on all three datasets.
PC-AVS gets slightly higher AVConf scores than our method on HDTF and Testset 1, but are much worse than ours on all the other metrics, especially LMD. 
This indicates that PC-AVS learns natural-looking lip movements but fails to capture individual speaking styles.
In contrast, our method is more personalized as it takes into account the identity parameters $z_{id}$ and thus excels in the more fine-grained LDM. 


\subsubsection{Qualitative Comparison}

\begin{figure}[t]
  \centering
   \includegraphics[width=\linewidth]{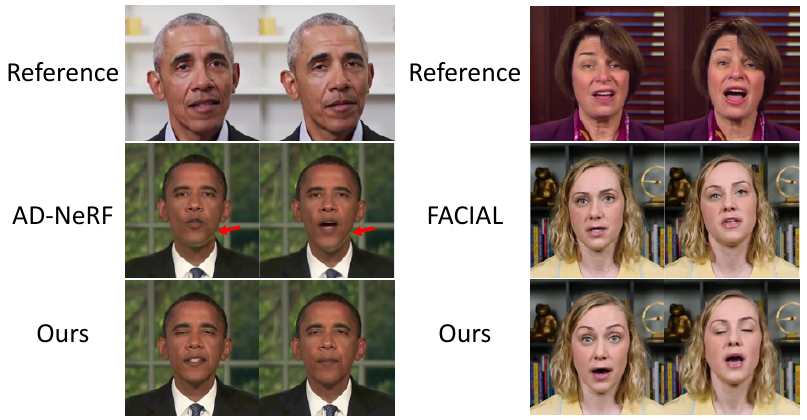}

   \caption{Comparison with AD-NeRF \cite{guo2021ad} and FACIAL \cite{zhang2021facial}. }
   \label{fig:3dmethod}
\end{figure}

We compare our methods with state-of-the-art 2D-based methods, including the explicit ATVG \cite{chen2019hierarchical} and MakeitTalk \cite{zhou2020makelttalk}, and implicit Wav2Lip \cite{prajwal2020lip} and PC-AVS \cite{zhou2021pose}, in \cref{fig:qualitative}. For 3D-based methods, we compare ours with the explicit FACIAL \cite{zhang2021facial} and implicit AD-NeRF \cite{guo2021ad} in \cref{fig:3dmethod}. 


As \cref{fig:qualitative} shows, our method produces the highest quality results with the most synchronized lip-movement. 
Specifically, i) ATVG and MakeitTalk fail to produce accurate lip movements as they rely on less expressive 2D facial landmarks; ii) 
Wav2Lip produces blurry mouths that do not match the sharp parts in the rest of the video frames, making the results unnatural; iii) although PC-AVS produces head movements that are consistent with the ground truth, it cannot well preserve the identity of the speaker. 
In addition, none of these methods can synthesize high-resolution videos.
In contrast, our method allows for the synthesis of high-resolution, high-quality videos with highly synchronized lip movements that preserve facial details well (\eg, teeth and wrinkles), which are crucial for identity preservation and the naturalness of facial reenactment.

\begin{table}
  \centering
  \begin{tabular}{lr|lr}
    \toprule
    Method & AVConf$\uparrow$ & Method & AVConf$\uparrow$ \\
    \cmidrule(r){1-2}\cmidrule(r){3-4}
    AD-NeRF \cite{guo2021ad} & 3.607 & FACIAL \cite{zhang2021facial} & 4.623 \\
    Ours & {\bf 6.758} & Ours & {\bf 6.678} \\
    \bottomrule
  \end{tabular}
  \caption{Quantitative comparisons of our method with AD-NeRF \cite{guo2021ad} and FACIAL \cite{zhang2021facial}.}
  \label{tab:diversity}
\end{table}

As \cref{fig:3dmethod} shows, our method is also superior to AD-NeRF and FACIAL.
Specifically, i) AD-NeRF suffers from the artifacts at the head-neck junction which stem from a mismatch between the two NeRFs it uses to model the head and torso separately;
ii) FACIAL produces less accurate lip movements due to the less expressive 3D face shape it uses as the intermediate face representation.
Please refer to~\cref{tab:diversity} for a quantitative comparison w.r.t lip movement accuracy between AD-NeRF, FACIAL and our method.

\vspace{2mm}
\noindent \textbf{Remark.}
For the comparison with AD-NeRF and FACIAL (\cref{fig:3dmethod}), we use the codes and pretrained models released by the authors and compare the generalizability by feeding all models with 5 unpaired and gender-balanced audio clips mentioned above. The reference images are the corresponding frames of the input audio clips in the new videos.

\begin{figure}[t]
  \centering
   \includegraphics[width=\linewidth]{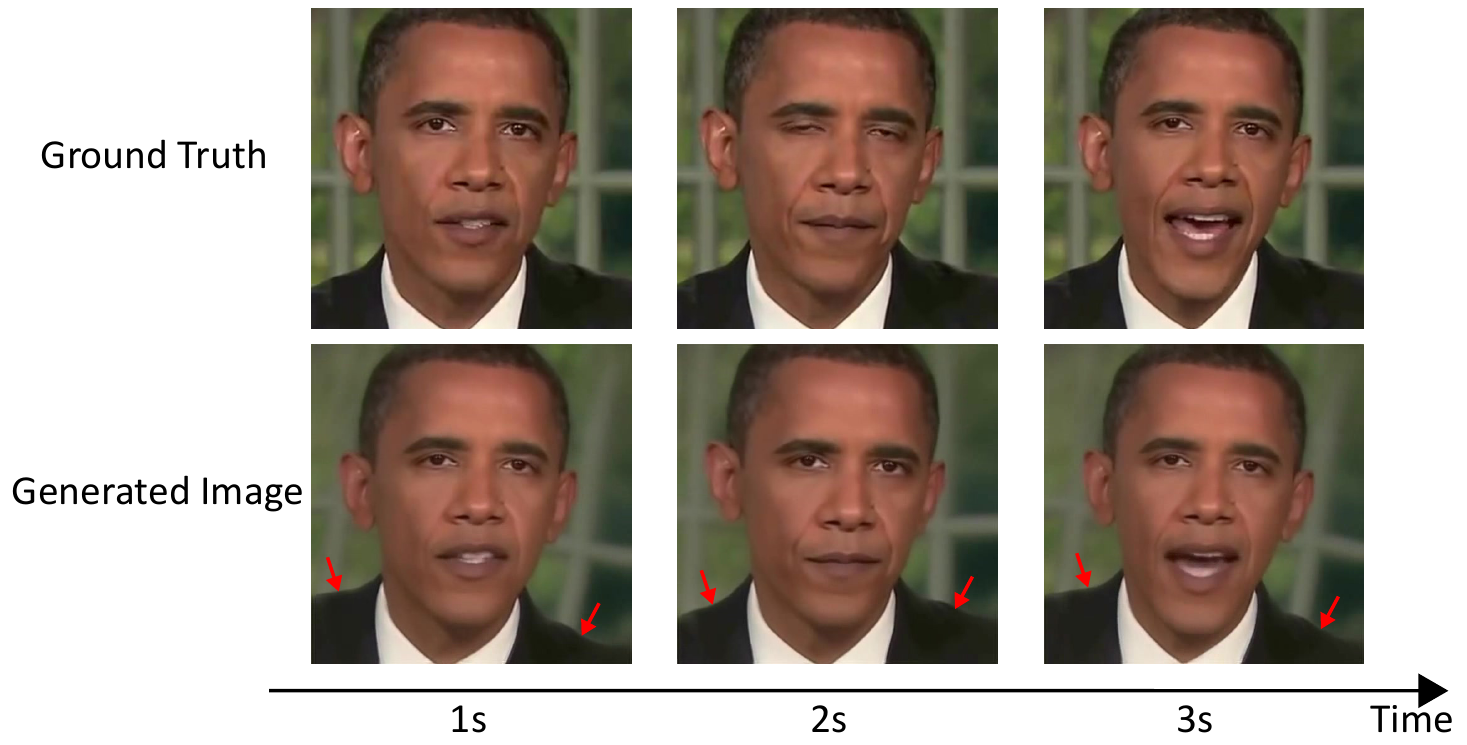}

   \caption{Ablation study of our rendering with PIR component. Row 2: the images generated without this component.}
   \label{fig:abl_first_branch}
\end{figure}

\begin{figure}[t]
  \centering
   \includegraphics[width=\linewidth]{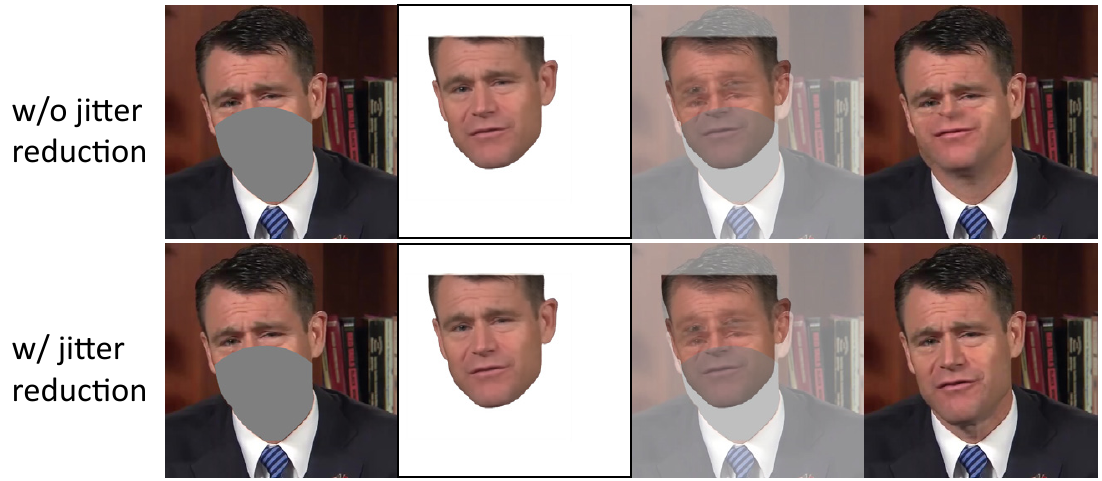}

\caption{Ablation study of our jitter reduction technique.}
   \label{fig:abl_intrinsic}
\end{figure}

\subsection{Ablation Study}

\noindent{\bf Rendering with PIR.} 
As \cref{fig:abl_first_branch} shows, without our rendering with PIR component (\ie, use the implicit representation parameterization component to generate the output image $I$ directly, not just the head), the torso changes rapidly within seconds and produces unnatural results. This justifies the necessity of our Rendering with PIR component.

\vspace{1mm}
\noindent{\bf Jitter Reduction.} As \cref{fig:abl_intrinsic} shows, without jitter reduction, the rendering component cannot align $I_F$ with $I_M$, thus producing unnatural videos.
This justifies the necessity of our jitter reduction technique. 


\begin{table}
  \centering
  \begin{tabular}{l|cccc}
    \toprule
    Method & Lip-sync & Image & Video \\
    \midrule
    ATVG \cite{chen2019hierarchical} & 2.87 & 1.87 & 1.76 \\
    Wav2Lip \cite{prajwal2020lip} & 3.91 & 2.67 & 2.76 \\
    MakeitTalk \cite{zhou2020makelttalk} & 2.69 & 2.73 & 2.84 \\
    PC-AVS \cite{zhou2021pose} & 3.89 & 3.16 & 3.38 \\
    AD-NeRF \cite{guo2021ad} & 3.84 & 3.96 & 3.73 \\
    FACIAL \cite{zhang2021facial} & 4.16 & 4.09 & 4.11 \\
    \midrule
    Ours & {\bf 4.31} & {\bf 4.20} & {\bf 4.29} \\
   
    \bottomrule
  \end{tabular}
  \caption{User study on lip-sync (audio-lip synchronization), image quality and video realness.}
  \label{tab:user_study}
\end{table} 

\subsection{User Study}

We invite 15 volunteers to participate in our user study to evaluate facial reenactment results based on three criteria: lip-sync (audio-lip synchronization), image quality and video realness. 
We create 3 videos for each method with the same audio input and ask the volunteers to give their ratings on a scale of 1 (worst) to 5 (best) for each video.
As \cref{tab:user_study} shows, our method scores the highest in all three criteria.


\section{Conclusion}
In this work, we propose an innovative facial reenactment framework based on our novel parametric implicit representation (PIR). 
Specifically, our PIR breaks the trade-off between interpretability and expressive power that plagued previous explicit and implicit methods, thus paving the way for controllable and high-quality audio-driven facial reenactment.
We have also devised several novel techniques to improve the three components of our framework. 
Extensive experiments demonstrate the effectiveness of our method.

\section*{Acknowledgments}
This work was supported in part by the Guangdong Basic and Applied Basic Research Foundation (NO.~2020B1515020048), in part by the National Natural Science Foundation of China (NO.~61976250), in part by the Shenzhen Science and Technology Program (NO.~JCYJ20220530141211024) and in part by the Fundamental Research Funds for the Central Universities under Grant 22lgqb25. This work was also sponsored by Tencent AI Lab Open Research Fund (NO.~Tencent AI Lab RBFR2022009).

{\small
\bibliographystyle{ieee_fullname}
\bibliography{egbib}
}

\end{document}


\title{Supplementary Material \\
Parametric Implicit Face Representation for Audio-Driven Facial Reenactment 
}


\author{Ricong Huang
$^{1}$ \quad Peiwen Lai$^{1}$ \quad Yipeng Qin$^{2}$ \quad Guanbin Li$^{1}$\thanks{Corresponding author is Guanbin Li.} \\
$^1${School of Computer Science and Engineering, Sun Yat-sen University} \quad $^2${Cardiff University} \\
{\tt\small \{huangrc3, laipw5\}@mail2.sysu.edu.cn, qiny16@cardiff.ac.uk, liguanbin@mail.sysu.edu.cn}
}

\maketitle

In this supplement, we provide more implementation details of the network architecture of the three components we proposed:
contextual audio to expression encoding (\cref{sec:audio2exp}), implicit representation parameterization (\cref{sec:eg3d}), rendering with parametric implicit representation (\cref{sec:rendering}). 
Please note that we also include an additional ablation study on the choice of hyper-parameter $k$ in \cref{sec:audio2exp}.
Lastly, we show additional qualitative evaluation results in \cref{sec:more_quali}.
We strongly encourage readers to watch our supplementary video, which demonstrates the superiority of our method. 

\section{Contextual Audio to Expression Encoding}
\label{sec:audio2exp}
\subsection{Network Architecture}
\label{sec:audio2exp_architecture}
\begin{figure}[t]
  \centering
   \includegraphics[width=\linewidth]{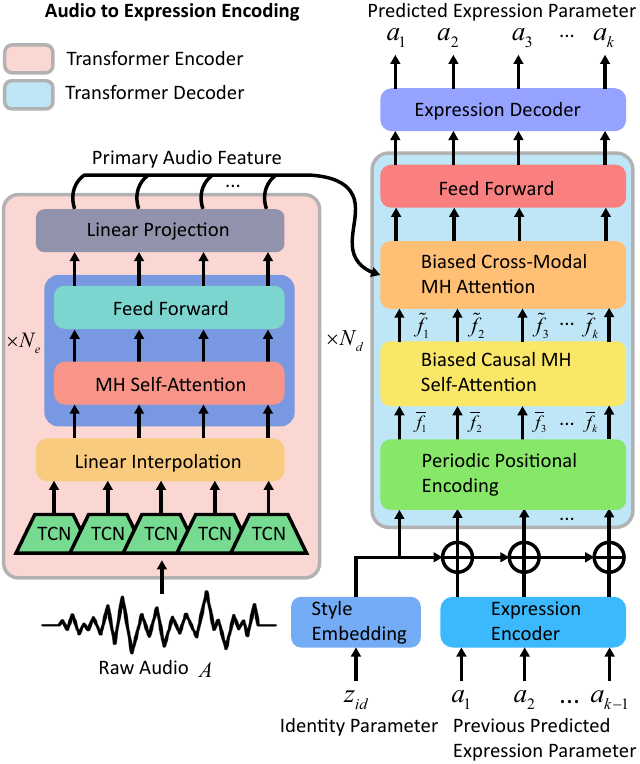}

   \caption{Network architecture of our contextual audio to expression encoding component.}
   \label{fig:audio2exp}
\end{figure}

As \cref{fig:audio2exp} shows, our contextual audio to expression encoding component is a transformer-based architecture similar to \cite{fan2022faceformer}, consisting of a transformer encoder and decoder. 

For the transformer encoder, we first extract the primary audio feature of a raw audio $A$ through wav2vec 2.0 \cite{baevski2020wav2vec}, which consists of an audio feature extractor and a multi-layer transformer encoder. Between them, the audio feature extractor consists of several temporal convolutions layers (TCN), and the transformer encoder is a stack of multi-head self-attention and feed-forward layers. 
Note that we have added a linear interpolation layer in-between to resample the audio features from the TCN output to ensure that they share the same sampling frequency with the training video.
The dimension of each block is 1024 and the number of attention heads is 16. 
A linear projection layer is added after the transformer blocks to project the extracted features to the input space of the biased cross-modal MH Attention blocks in the transformer decoder.

The transformer decoder takes input from the output of the transformer encoder, the style embedding layer, and the expression encoder. Its output is converted by the expression decoder into the predicted expression parameter $a_1, a_2, ..., a_k$. We use a sequence of $k=100$ video frames for training. 
Among them, the style embedding and expression encoder are both fully-connected layers with a dimension of 1024; the expression decoder is a fully-connected layer with a dimension of 64.
The style embedding layer takes the identity parameter $z_{id}$ as input. The expression encoder takes the previous predicted expression parameter $a_1, a_2, ..., a_{k-1}$ as input.
Their outputs are summed together except for the first frame when only $z_{id}$ is available. 
Like \cite{fan2022faceformer}, each block in the transformer decoder consists of a periodic positional encoding layer, a biased causal multi-head self-attention layer, a biased cross-modal multi-head attention layer and a feed forward layer, whose details are described as follows.


\vspace{2mm}
\noindent \textbf{Periodic Positional Encoding (PPE).}
PPE is used for the injection of temporal order and is formulated as:
\begin{equation}
    \begin{aligned}
    PPE_{(t, 2i)} &= \sin\left((t\mod p) / 10000^{2i/d}\right) \\
    PPE_{(t, 2i+1)} &= \cos\left((t\mod p) / 10000^{2i/d}\right)
    \end{aligned},
\end{equation}
where $p=25$ indices, the period $t$ denotes the current time-step in the input sequence, $d$ is its dimension and $i$ is the dimension index. 
The output of PPE is a sequence of features $\overline{F}_t=(\overline{f}_1,\dots,\overline{f}_t)$, $1 \leq t \leq k$.

\vspace{2mm}
\noindent \textbf{Biased Causal Multi-head (MH) Self-attention.}
This layer is designed to ensure causality and to improve the generalization of the model to long sequences, which is formulated as:
\begin{equation}
    \begin{aligned}
    &{\rm MH}(Q^{\overline{F}},K^{\overline{F}},V^{\overline{F}},B^{\overline{F}}) = {\rm Concat}({\rm head_1}, \dots,{\rm head_{H}})W^{\overline{F}} \\
    &{\rm head_h} = {\rm softmax}\left(\frac{Q^{\overline{F}}_h(K^{\overline{F}}_h)^T}{\sqrt{d_k}}+B^{\overline{F}}_h\right)V^{\overline{F}}_h \\
    &B^{\overline{F}}(i,j) = \begin{cases}
        \lfloor (i-j)/p \rfloor,\quad & j \leq i \\
        -\infty,\quad & {\rm otherwise}
        \end{cases} \\
    &B^{\overline{F}}_h = B^{\overline{F}}m
    \end{aligned},
\end{equation}
where $Q^{\overline{F}},K^{\overline{F}},V^{\overline{F}}$ are projected from the sequence $\overline{F}_t=(\overline{f}_1,\dots,\overline{f}_t)$, $W^{\overline{F}}$ is a parameter matrix, $d_k$ is the dimension of $Q^{\overline{F}}$ and $K^{\overline{F}}$, $B^{\overline{F}}$ is the temporal bias matrix and $i, j$ are the indices of it, and $m$ is a head-specific slope. 
The output of this layer is a sequence of features $\tilde{F}_t=(\tilde{f}_1,\dots,\tilde{f}_t)$, $1 \leq t \leq k$.

\vspace{2mm}
\noindent \textbf{Biased Cross-modal Multi-head (MH) Attention.}
This layer combines the output of the transformer encoder and $\tilde{F}_t=(\tilde{f}_1,\dots,\tilde{f}_t)$, which is formulated as:
\begin{equation}
    \begin{aligned}
    &{\rm Att}(Q^{\tilde{F}},K^{A},V^{A},B^{A}) = {\rm softmax}\left(\frac{Q^{\tilde{F}}(K^{A})^T}{\sqrt{d_k}}+B^{A}\right)V^{A} \\
    &B^{A}(i,j) = \begin{cases}
        0,\quad & i \leq j < (i+1) \\
        -\infty,\quad & {\rm otherwise}
        \end{cases}
    \end{aligned},
    \label{equ:cross-modal}
\end{equation}
where $B^{A}$ is the alignment bias matrix. \Cref{equ:cross-modal} is extended to $H$ heads to explore different subspaces. 

\vspace{2mm}
\noindent \textbf{Feed Forward Layer.} A fully-connected layer of dimension 2048.

For both the biased causal MH self-attention and the biased cross-modal MH attention, 4 heads are employed with a model dimension of 1024. 
We use $N_e=24$ and $N_d=1$ transformer blocks in our implementation.


\subsection{Ablation Study on Sequence Length $k$}
As mentioned in our main paper, our audio to expression encoding is a stand-alone and light-weight task that can learn the contextual information from {\it long} audio sequences. 
To justify the benefit brought by long sequences, we conduct an ablation study on the sequence length $k$.
As \cref{tab:k} shows, in general, the generated talking head videos achieve better LMD and AVConf scores with longer training audio sequences, indicating that capturing the long-term contextual information from audio sequences helps generate highly synchronized lip movements for talking portraits.
We use $k=100$ in our method as it achieves the best scores.

\begin{table}
  \centering
  \begin{tabular}{l|ccccc}
    \toprule
    k & 1 & 10 & 50 & 100 & 200 \\
    \midrule
    LMD$\downarrow$ & 2.556 & 2.273 & 1.626 & {\bf 1.477} & 1.650 \\
    AVConf$\uparrow$ & 4.201 & 4.505 & 6.873 & {\bf 7.071} & 6.929 \\
    \bottomrule
  \end{tabular}
  \caption{Ablation study of hyper-parameter $k$ (sequence length). $k=1$ indicates no use of contextual information.}
  \label{tab:k}
\end{table} 

\section{Implicit Representation Parameterization}
\label{sec:eg3d}
We implement our implicit representation parameterization based on an efficient tri-plane structure \cite{chan2022efficient}. 

As shown in Fig. 2 of the main paper, we employ a mapping network to produce a 512-D intermediate latent vector $z$ from the concatenation of identity parameter $z_{id}$ and expression parameter $z_{exp}$. Conditioned on the latent vector $z$, a StyleGAN2 \cite{karras2020analyzing} generator is employed to generate a $96\times 256\times 256$ feature map. 
Then the feature map is split into three axis-aligned orthogonal feature planes, each with a resolution of $32\times 256\times 256$. 
Given camera pose $R, t$ and intrinsic matrix $K$, a 3D point position in world coordinates $[x_w\;y_w\;z_w\;1]^T$ is calculated based on:
\begin{equation}
    z_c\begin{bmatrix}
    u \\ 
    v \\
    1
    \end{bmatrix} = K \begin{bmatrix}
    R & t \\ 
    \mathbf{0} & 1
    \end{bmatrix}\begin{bmatrix}
    x_w \\ 
    y_w \\
    z_w \\
    1
    \end{bmatrix},
\end{equation}
where $[u\;v\;1]^T$ is the 2D point position in pixel coordinates and $z_c$ is the z-coordinate of 3D point position in camera coordinates. 
Then we project it onto each of the three feature planes, retrieve the feature vector $(F_{xy}, F_{xz}, F_{yz}$ via bilinear interpolation, and aggregate the three feature vectors via summation. These aggregated features are interpreted as 32-D color and 1-D density through a lightweight decoder which is a multi-layer perceptron with a single hidden layer and a softplus activation function. Furthermore, they are reconstructed as a $32\times 64\times 64$ feature map $I_F$ using volume rendering. Two-pass importance sampling strategy is used to implement volume rendering \cite{max1995optical} as in \cite{mildenhall2021nerf}. 
Compared to NeRF structures using large fully connected networks \cite{mildenhall2021nerf}, the computational cost of tri-plane based neural rendering is reduced since it has a smaller decoder.
\begin{figure*}[t]
  \centering
   \includegraphics[width=\linewidth]{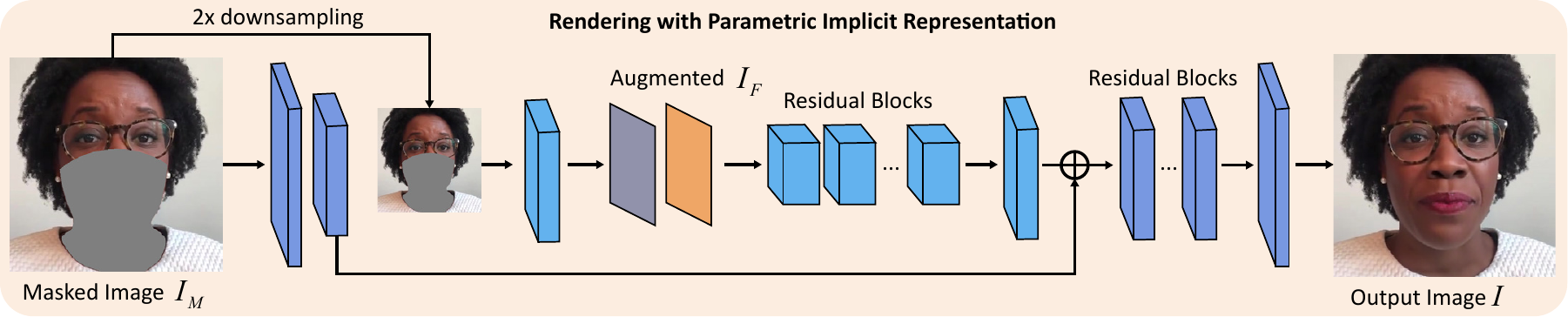}

  \caption{Network architecture of our rendering with parametric implicit representation component. We formulate facial reenactment as an image inpainting problem conditioned on the implicit representation $I_F$ and use a coarse-to-fine network structure to tackle it. }
   \label{fig:rendering}
\end{figure*}
\begin{figure*}[t]
  \centering
  \includegraphics[width=\linewidth]{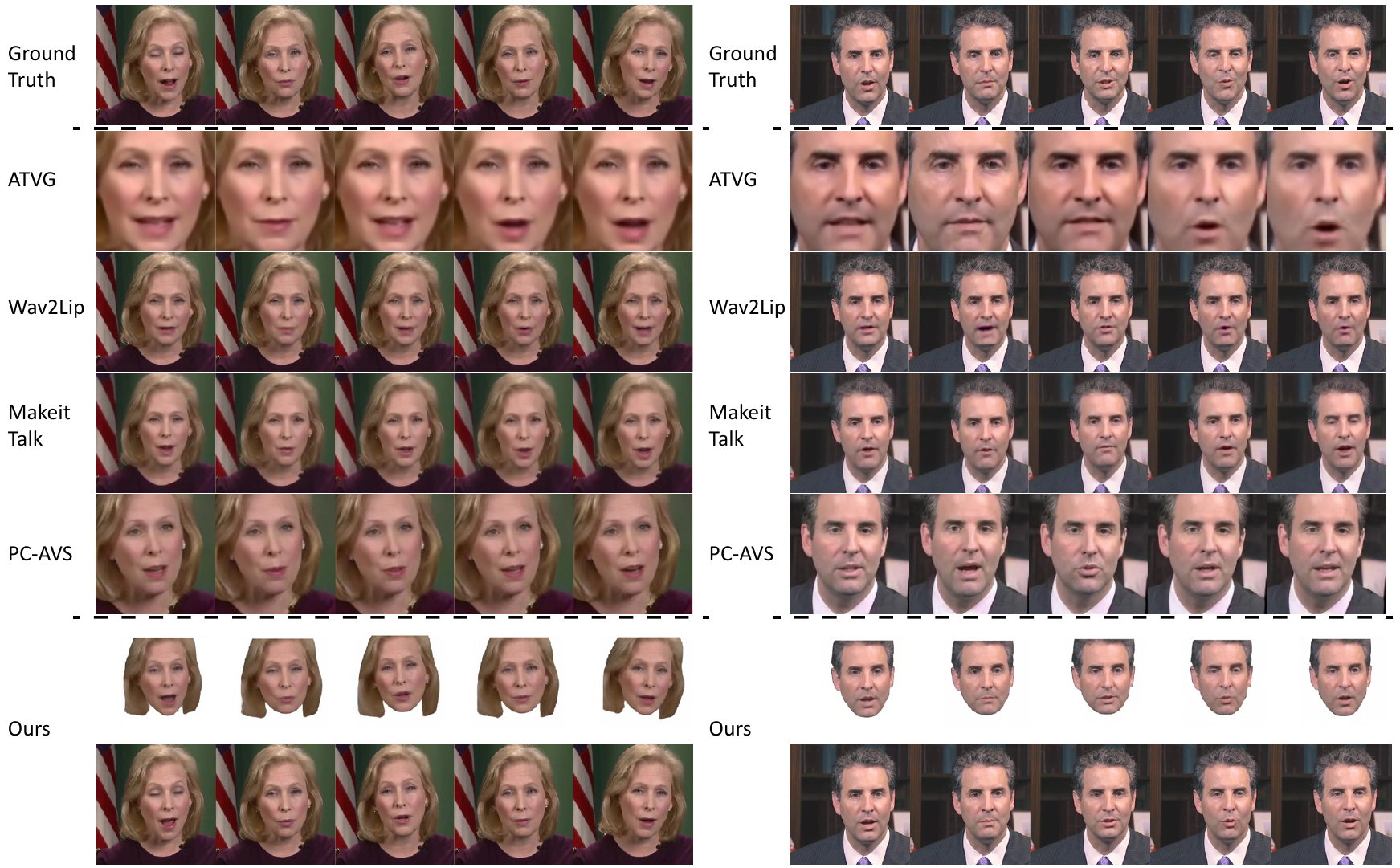}

\caption{Additional qualitative comparison with ATVG \cite{chen2019hierarchical}, Wav2Lip \cite{prajwal2020lip}, MakeitTalk \cite{zhou2020makelttalk} and PC-AVS \cite{zhou2021pose}. 
}
   \label{fig:sup_qualitative}
\end{figure*}
Finally, a CNN-based upsampling network is used to upsample and render $I_F$ to the final image $I_{face}$ with a resolution of $3\times 512\times 512$. 


\section{Rendering with PIR}
\label{sec:rendering}
As shown in \cref{fig:rendering}, our rendering with parametric implicit representation (PIR) component uses a coarse-to-fine generator \cite{wang2018high} to generate the output image. The input $3\times 512\times 512$ masked image $I_M$ is downsampled to a $3\times 256\times 256$ image through the average pooling. Then the image is further processed through several convolution layers and concatenated with the augmented feature map $I_F$ from the implicit representation parameterization component. The concatenated features are further processed through 4 residual blocks and several convolution layers. After that, the features is summed with the feature maps extracted from $I_M$, passing through 3 residual blocks and finally rendering into the output image $I$ through convolution layers. 

\begin{figure*}[t]
  \centering
  \includegraphics[width=\linewidth]{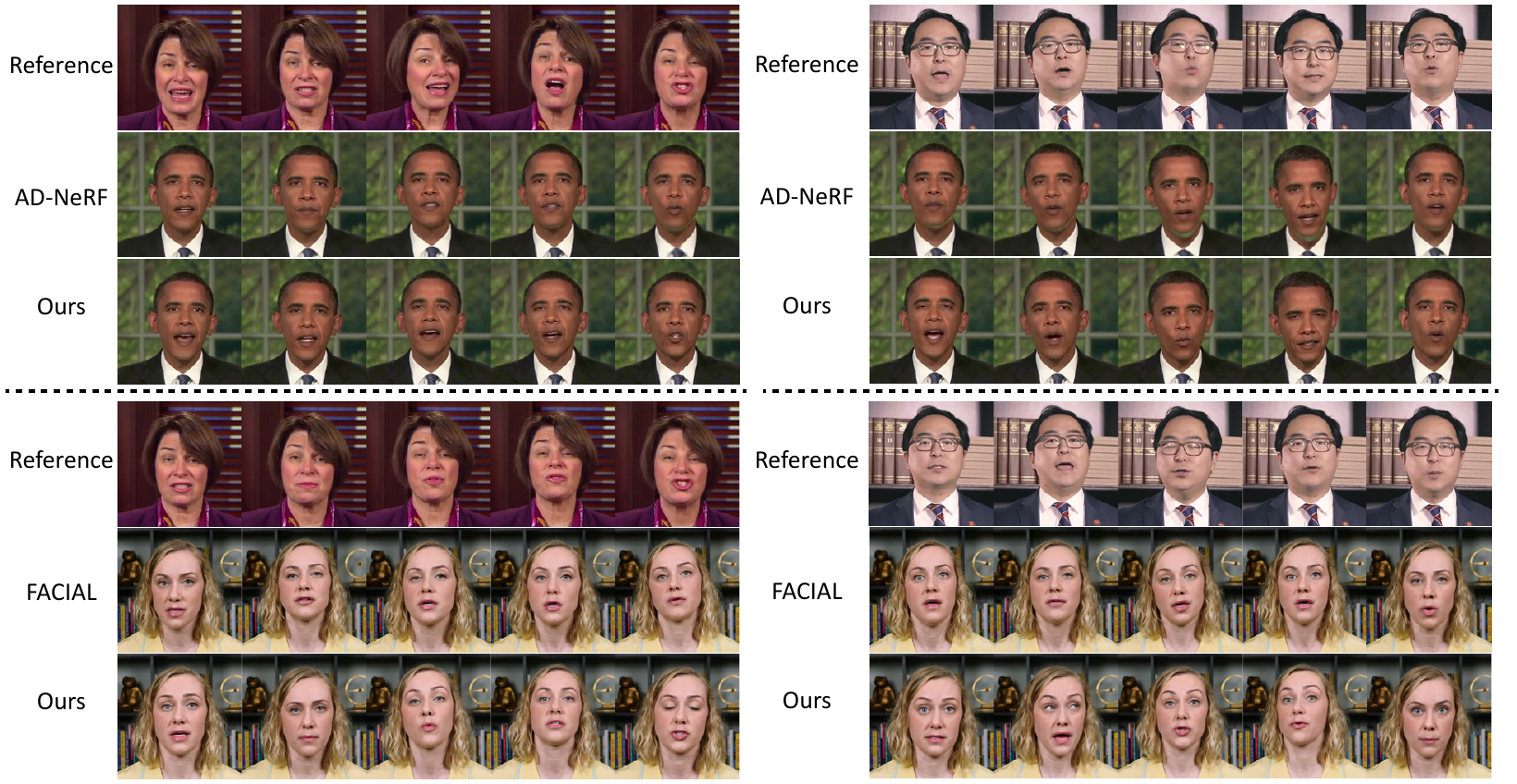}

  \caption{Comparison with AD-NeRF \cite{guo2021ad} and FACIAL \cite{zhang2021facial}.}
  \label{fig:sup_3dmethod}
\end{figure*}

\section{Additional Qualitative Results}
\label{sec:more_quali}
As a complement to the main paper, we show additional qualitative results of ATVG \cite{chen2019hierarchical}, Wav2Lip \cite{prajwal2020lip}, MakeitTalk \cite{zhou2020makelttalk}, PC-AVS \cite{zhou2021pose} and our method. As \cref{fig:sup_qualitative} shows, it can be observed that our method generates talking portraits with highly synchronized lip movements and high fidelity to the facial details, outperforming all previous methods. 

We also show more qualitative comparison results with AD-NeRF \cite{guo2021ad} and FACIAL \cite{zhang2021facial} in \cref{fig:sup_3dmethod}. The generated talking portraits are driven by the audio from different identities.
The results show that the talking heads generated by AD-NeRF \cite{guo2021ad} have obvious artifacts at the head-neck junction, and those generated by FACIAL \cite{zhang2021facial} have less accurate lip movements. 
In contrast, our method can generate natural and vivid talking portraits, indicating that it generalizes better to unseen audios.


{\small
\bibliographystyle{ieee_fullname}
\bibliography{egbib}
}